\documentclass[runningheads]{llncs}
\usepackage[T1]{fontenc}
\usepackage{graphicx}
\usepackage{algorithm}
\usepackage{algorithmic}
\usepackage{booktabs}
\usepackage[misc]{ifsym}
\usepackage{mathtools}

\usepackage{mwe}

\begin{document}

\title{Attending To Syntactic Information In Biomedical Event Extraction Via Graph Neural Networks}

\author{Farshad Noravesh\textsuperscript{1} \and
Reza Haffari\textsuperscript{2} \and
Ong Huey Fang\textsuperscript{3} \and
Layki Soon\textsuperscript{4} \and 
Sailaja Rajalana\textsuperscript{5}  \and
Arghya Pal\textsuperscript{6}
}

\institute{
Monash University, Malaysia \email{Farshad.Noravesh@monash.edu}
\and
Monash University, Australia \email{Gholamreza.Haffari@monash.edu}
\and
Monash University, Malaysia \email{ong.hueyfang@monash.edu}
\and
Monash University, Malaysia \email{soon.layki@monash.edu}
\and
Monash University, Malaysia \email{Sailaja.Rajanala@monash.edu}
\and
Monash University, Malaysia \email{arghya.pal@monash.edu}
}
   
\maketitle              

\begin{abstract}
Many models are proposed in the literature on biomedical event extraction(BEE). Some of them use the shortest dependency path(SDP) information to represent the argument classification task. There is an issue with this representation since even missing one word from the dependency parsing graph may totally change the final prediction. To this end, the full adjacency matrix of the dependency graph is used to embed individual tokens using a graph convolutional network(GCN). An ablation study is also done to show the effect of the dependency graph on the overall performance. The results show a significant improvement when dependency graph information is used. The proposed model slightly outperforms state-of-the-art models on BEE over different datasets.
\end{abstract}

\section{introduction}

\begin{figure}
\centering 
\includegraphics[scale=0.40]{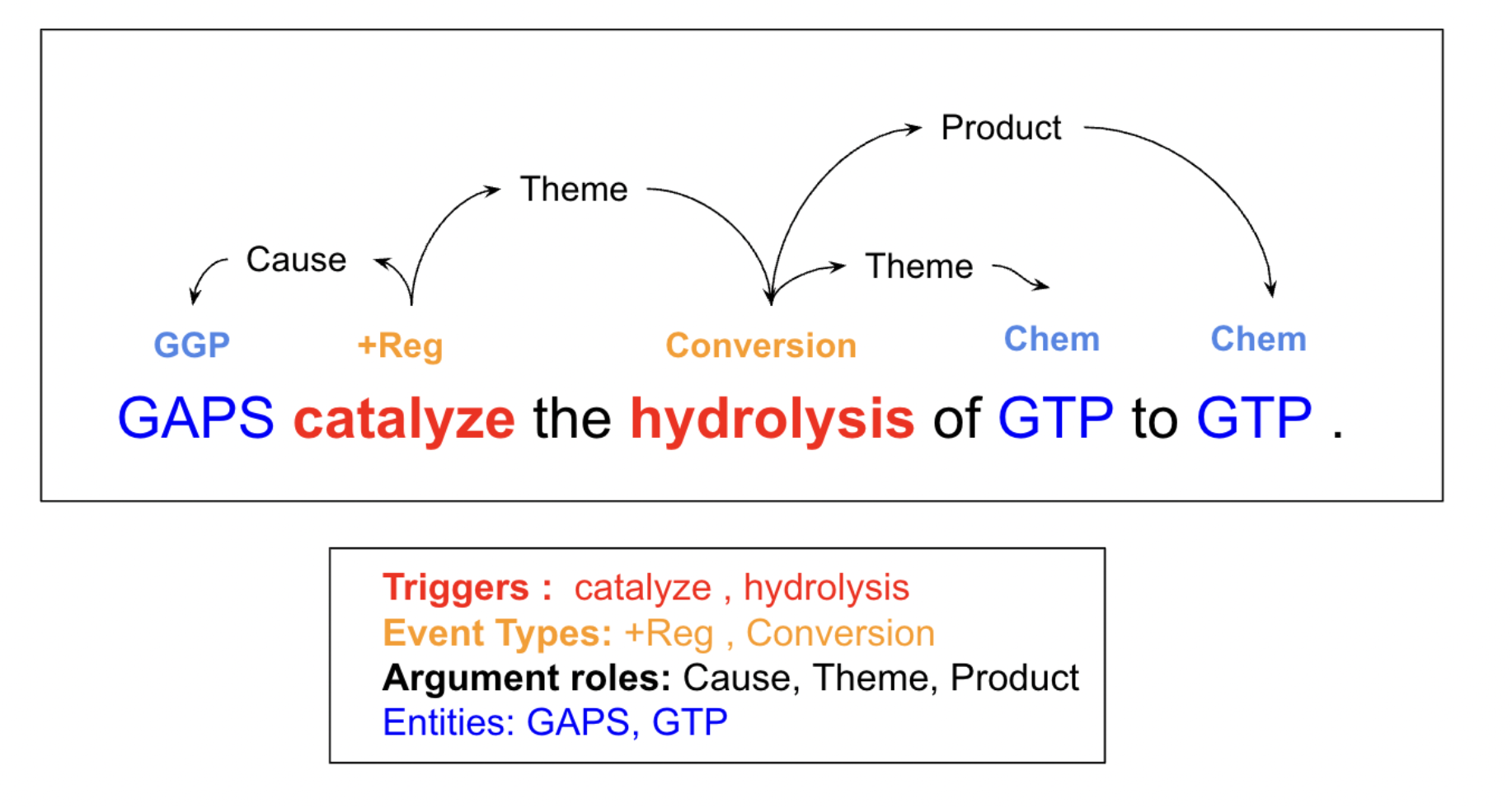}
\caption{An example of pathway curation sample from BioNLP13-pc}
\label{fig-bioNLP13-pc}
\end{figure}

BEE is a structured prediction task since all nodes and edges of the graph should be predicted simultaneously. Since the complex nature of graph could not be linearized, the performance of LLM on BEE is very poor and suffers from hallucination and explainability issues.
An example of BEE from pathway curation subtask (pc) of  BioNLP13 dataset is shown in Figure~\ref{fig-bioNLP13-pc} which is an example of a nested event structure. A trigger may have an argument which is itself a trigger for another event. Dependency parsing provided rich information to help the model explain different relations specially when the distance between trigger and argument is long.

\par
In a dependency parse tree, the words of a sentence are represented as nodes, and the relationships between them are represented as labeled arcs. Each word in the sentence is assigned a syntactic role, such as a subject, object, or modifier, and the arcs indicate the direction and nature of the dependency relationship between the words. For example, a verb might have an arc pointing to its subject, indicating the subject is dependent on the verb.
There exists many open source dependency parsing libraries such as Scispacy and StanfordNLP which provide very accurate dependency parsing graph. Nonetheless, some researchers argue that these models have some errors that propagates downstream task and they prefer to use large language models(LLM) to circumvent this issue but LLM has not source of syntactic information. 
Although these tools have small error that may propagate downstream tasks, they are the only source of dependency information that could reveal rich syntactic information and the overall gain we achieve from them is higher than the small dependency parsing modeling error.

\begin{figure}[t!]
  \centering
    \includegraphics[width=0.9\textwidth]{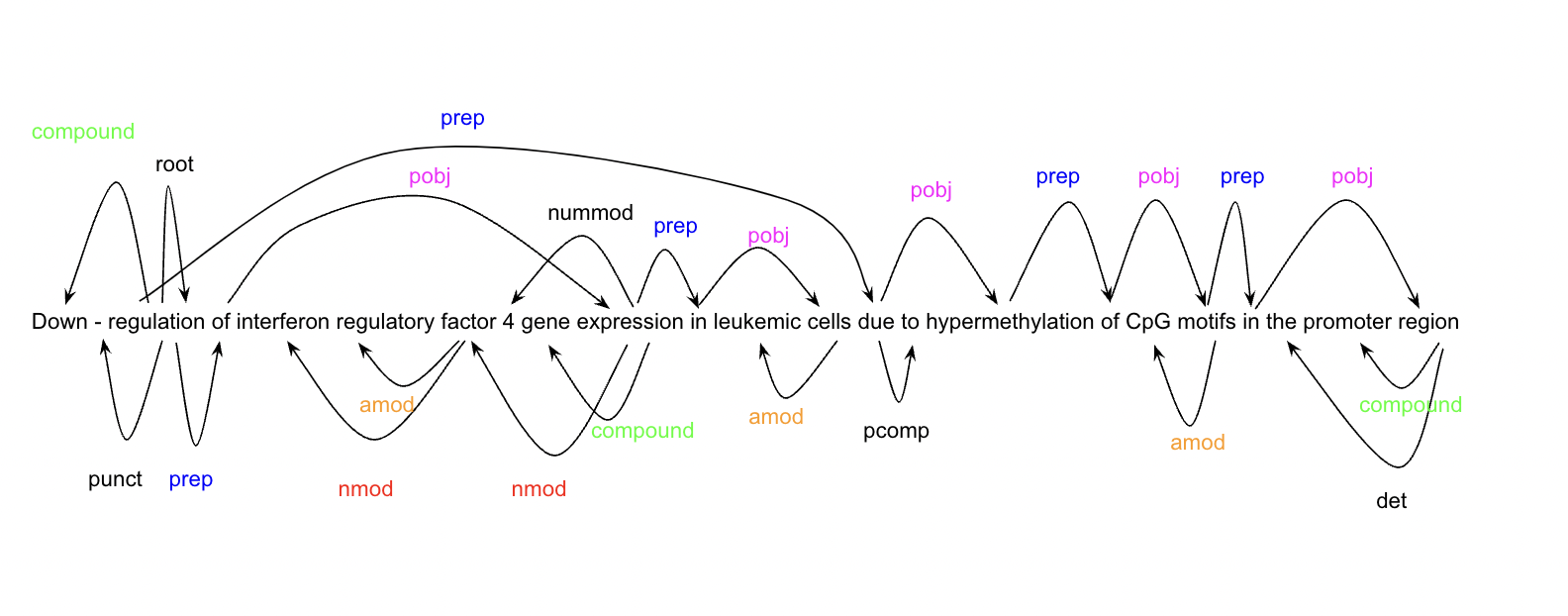}
  \caption{an example of a dependency parse from BioNLP2011}
  \label{fig-dependencyParse_bioNLP2011}
\end{figure}

Scispacy is leveraged for the present work to get the dependency parsing information and thus no learning is required to obtain such syntactic information. Figure~\ref{fig-dependencyParse_bioNLP2011} shows an example of a dependency parse that is obtained from Scispacy by applying it to a sentence in BioNLP2011 dataset.

\begin{figure}[H]
\centering 
\includegraphics[scale=0.40]{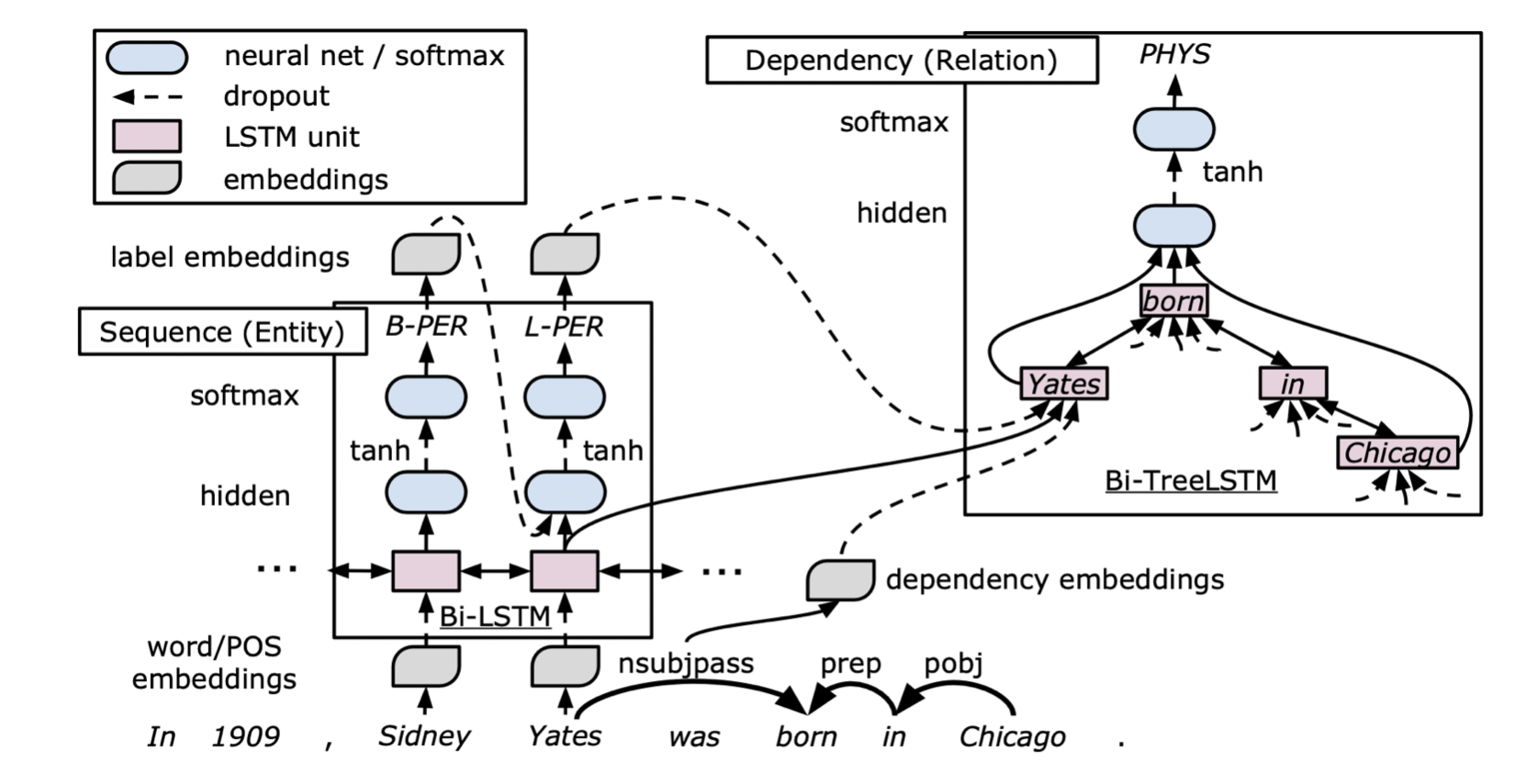}
\caption{The architecture of tree LSTMRNN model proposed in \cite{MakotoMiwa2016} }
\label{fig-twosubpath}
\end{figure}

\begin{figure}
\centering 
\includegraphics[scale=0.60]{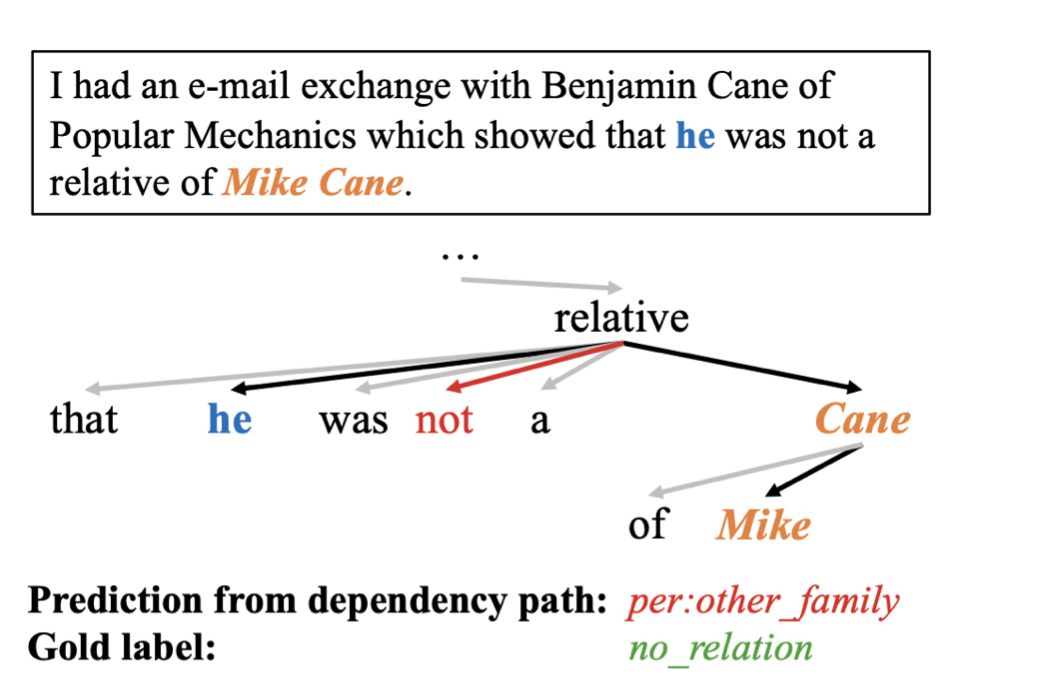}
\caption{taken from \cite{YuhaoZhang2018} that shows SDP between two entities in the bold. }
\label{fig-subtreeManning}
\end{figure}

In the BEE literature, each research paper only leverages part of dependency parsing graph and they neglect the rich information in the  full graph. For example, \cite{MakotoMiwa2016} uses shortest path between a pair of target words in the given dependency parsing graph. In particular it creates two subpath between a pair of target words. The first subpath is from the first entity to the least common node and the second subpath is from the least common node to the second entity. These two path are effective in relation extraction between any two entities. With the same analogy this idea could be used for other tasks such as event extraction to find subpaths between trigger and another trigger or between a trigger to an entity. Figure~\ref{fig-twosubpath} shows how these two subpath could be used in RE.
\cite{YuhaoZhang2018} uses GNN for better representations of entities for the task of relation extraction but the present paper aims to extend the use of GNN to BEE. \cite{YuhaoZhang2018} introduced a contextualized GCN (C-GCN) model, where the input word vectors are first fed into a BiLSTM network to generate contextualized representations, which are then used as to initialise the GNN.
Pruning too aggressively by keeping only the dependency path could lead to loss of crucial information and conversely hurt robustness as  \cite{YuhaoZhang2018} showed in a simple example in Figure~\ref{fig-subtreeManning} that sometimes missing one word like "not" could change the final prediction completely and therefore using SDP between two entities may produce a nonexpressive representation.
To address this issue, \cite{YuhaoZhang2018} proposed path-centric pruning, a novel technique to incorporate information off the dependency path. This is achieved by including tokens that are up to distance K away from the dependency path in the
lowest common ancestor (LCA) subtree. $K = 0$, corresponds to pruning the tree down to the path, $K = 1$ keeps all nodes that are directly attached to the path, and $K =  \infty$ retains the entire LCA subtree. They combine this pruning strategy with GCN model, by directly feeding the pruned trees into the graph convolutional layers.They showed that pruning with $K = 1$ achieves the best balance between including relevant information (e.g., negation and conjunction) and keeping irrelevant content out of the resulting pruned tree as much as possible. 
\par
The present paper solves the four subtasks of BEE using contextual representation of BioBert as well as appropriate embedding using GNN. The model augments trigger and argument networks after the GNN layers to reduce the effect of oversmoothing, oversquashing and lack of appropriate positional encoding that are critical in any deep GNN. 

The following are four major contributions of the present paper:
\begin{enumerate}
\item{To create two models that have the same architecture but one has used dependency parsing graph and the other one does not leverage dependency parsing information. The effect of this added information is analysed experimentally.}
\item{To use GNN as an embedding on top of the contextual representation of BioBert which provides better representation of individual tokens in each sentence.}
\item{The architecture is designed in such a way that eliminates poor expressivity of graph convolutional networks. The MLP networks for head(trigger) and dependent(argument) has the potential to reduce poor modelling of graph convolutional networks.}
\item{The attention of BEE subtasks is on full dependency parsing graph that is implemented by GNN in contrast to other research papers that eliminate sensitive tokens that lies in the subgraph connecting trigger and their arguments.}
\end{enumerate}
\section{Modeling}
\subsection{Data Preprocessing}
Four datasets has been used in the present work and are freely available in Huggingface website which separates training,validation and test data for all datasets. The first dataset is Genia11 which is a subtask of BioNLP11. The next three datasets are three subtasks of BioNLP13 namely genia13, path curation(pc), and cancer genetics(cg).
At preprocessing step, the arguments, triggers as well as event structure that shows how trigger-argument pairs are linked together. Since nested events are of interest, a trigger could be connected to another trigger to create a nested structure. Please note that each argument could be either an entity or a trigger for another event.
\subsection{BioBertBEE And GNNBEE}
\begin{figure}
\centering 
\includegraphics[scale=0.50]{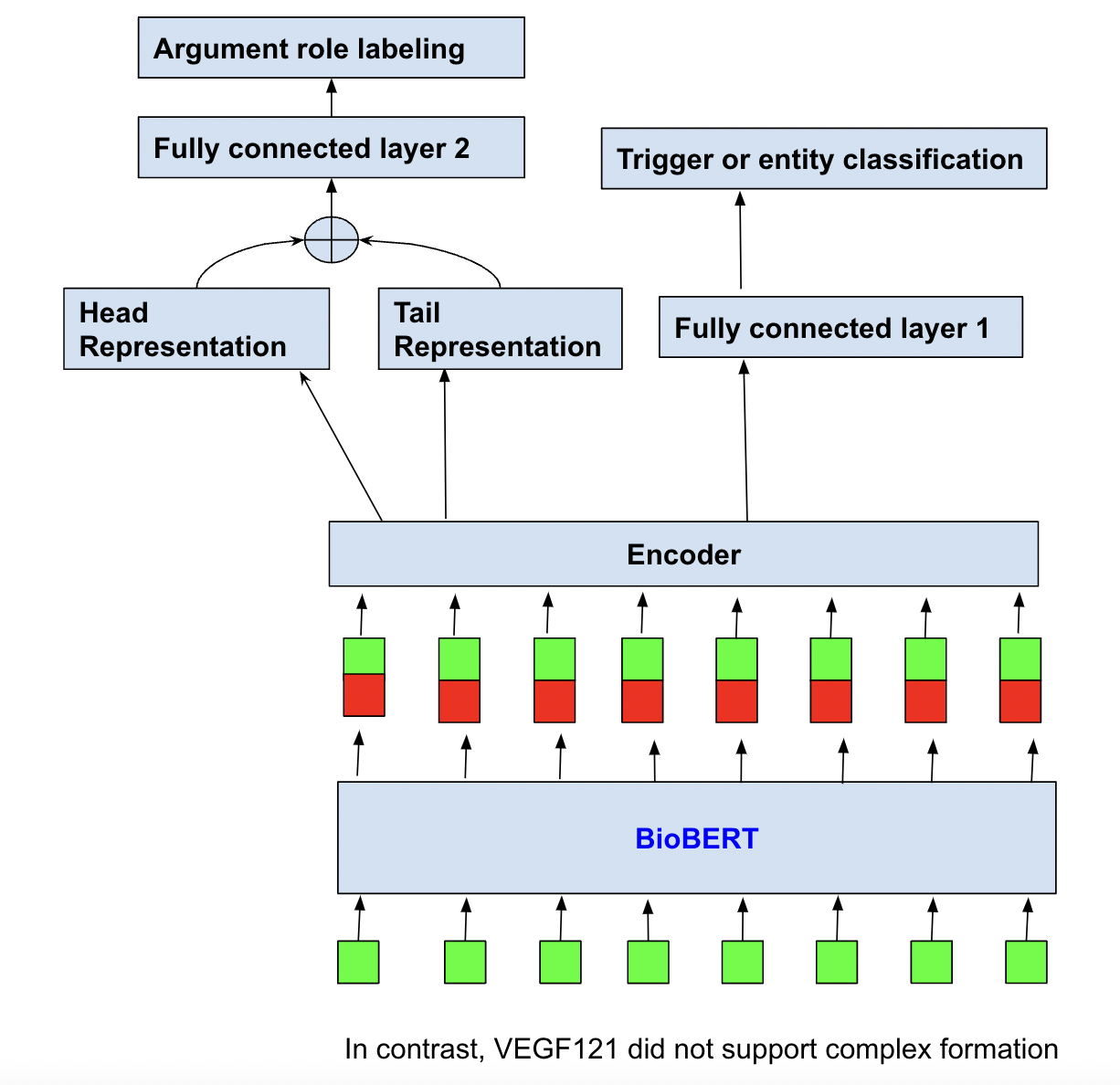}
\caption{architecture of bioBert based biomedical event extraction model}
\label{fig-bioBertBEE}
\end{figure}

\begin{figure}
\centering 
\includegraphics[scale=0.50]{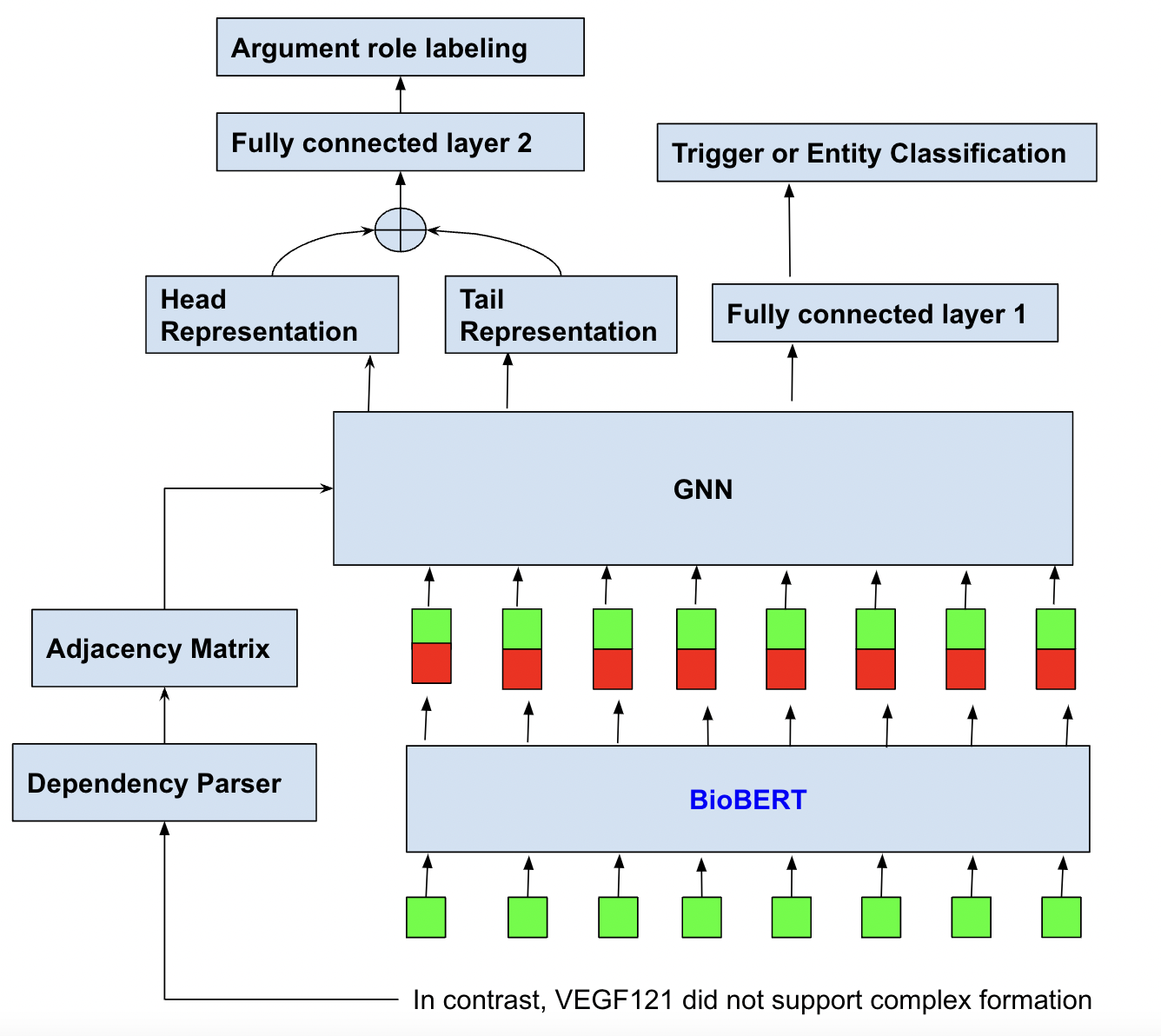}
\caption{architecture of GNN based biomedical event extraction model which is initialised by the encoder of bioBertBEE}
\label{fig-GNNBEE}
\end{figure}

GNN is very sensitive to initialisation and there exists some papers that focus only on this initialisation such as \cite{Han2022} and \cite{Abboud2021} that show how MLP architecture could be used for better training of GNN and consider MLP as a special case of GNN.
The present work randomly initialises the GNN. The BioBertBEE architecture is illustrated in Figure~\ref{fig-bioBertBEE}. In the argument role classification stage, each possible pair of trigger and entity mention is mixed and labeled with corresponding argument role. 
The weights of GNN feature vectors for each token is now initialised by the encoder outputs of the bioBertBEE. Other blocks such as trigger representation and entity representation is also initialised by the the corresponding blocks in the bioBertBEE as show in Figure~\ref{fig-GNNBEE}.
\subsubsection{Entity Or Trigger Classification}
One could treat both entity and trigger classification as a sequence labeling problem such as \cite{Zhang2021} that drew inspiration from conditional random field (CRF). To this end, similar to \cite{Zhang2021},  the previous tag embedding($s_{i-1}$) with the current tag embedding($s_{i}$) are concatenated. In the present work, the simplest sequence tagging model is used as follows:
\begin{equation}
P(s_{i}) = \text{Softmax}(\text{RELU}(W_{s}*(h_{i})+b_{s}))
\end{equation}
\subsubsection{Argument Role Classification}
To be able to efficiently model argument role labeling in BEE, features of trigger and entity should be combined with each other. Combining trigger and entity could be done in the following general approach which is first introduced in \cite{Dozat2017} and is implemented in \cite{JiTao2019} as follows:
\begin{equation}
\begin{split}
h_i &= \text{MLP}_{h}(c_{i}) \\
d_i &= \text{MLP}_{d}(c_{i}) \\
\sigma(i,j)&= \text{Softmax}_{i}(h_{i}^{T}Ad_{j}+b^{T}h_{i}+b^{T}d_{j})
\end{split}
\end{equation}
where $h_{i}$ and $d_{i}$ denote the head and dependent tokens. Thus, unlike  \cite{Huang2020} that uses only one network for trigger, the present work uses two separates network for trigger and argument respectively. There is a natural analogy between head and dependent token in dependency parsing with the trigger and argument in BEE. The ability to model the data structure of nested events is another reason to use the words head and dependent instead of trigger and entity which resembles a nested event structure as a graph.
To reduce the model complexity and the number of parameters, the following model is used for numerical experiments in the present work:
\begin{equation}\label{eq-concat}
\sigma(i,j) = \text{Softmax} (\text{RELU}(W_{r}*(h_{i}|d_{j})+b_{r}))
\end{equation}
where | in Equation~\ref{eq-concat} is the concatenation of $h_{i}$, $d_{j}$, $W_{r}$ and $b_{r}$ are the parameters of a linear network.
It should be mentioned that modeling the argument role labeling in Equation~\ref{eq-concat} is similar to \cite{Huang2020} since it first concatenates the i-th and j-th token. 
\subsubsection{Tackling Oversmoothing With MLP}
Oversmoothing is a fundamental problem in deep GNN which occurs when the stacked aggregators would make node representations converge to indistinguishable vectors. Oversmoothing increases as the number of layers of GNN increases. In the present work, a simple approach to tackle oversmoothing is designed based on two MLP networks that simulates k-hop distances without even using k layers in GNN. The MLP encodes long distance relationships between distant nodes and establishes a new embedding based on embedding of GNN. 
\subsubsection{Training}
The first proposed model does not use dependency parsing information and is trained using Algorithm~\ref{alg:BioBert-BEE-Model}.
\begin{algorithm}
  \begin{algorithmic}
    \STATE Input : (sentence, event structure) from dataset \\
    \STATE 1: preprocess the data \\
    \STATE 3: unfreeze the last layer of BioBert \\
    \STATE 4: Train BioBertBEE  \\
    \STATE 5: postprocess the predictions to evaluate TI,TC,AI,AC \\
    \STATE Output: performance measure at each epoch
    \caption{BioBert-BEE algorithm}
    \label{alg:BioBert-BEE-Model}
  \end{algorithmic}
\end{algorithm}
The second proposed model leverages full dependency parsing graph information and is trained using Algorithm~\ref{alg:bioBert-GNN-Model}.
\begin{algorithm}
  \begin{algorithmic}
    \STATE Input : (sentence, event structure) from dataset \\
    \STATE 1: preprocess the data \\
    \STATE 2: normalise the full adjacency matrix
    \STATE 3: unfreeze the last layer of BioBert \\
    \STATE 4: Train BioBert-GNN-BEE  \\
    \STATE 5: postprocess the predictions to evaluate TI,TC,AI,AC \\
    \STATE Output: performance measure at each epoch
    \caption{BioBert-GNN-BEE algorithm}
    \label{alg:bioBert-GNN-Model}
  \end{algorithmic}
\end{algorithm}
\section{Experiments}
\subsection{Evaluation}
The performance measures of the four subtasks of BEE are defined as follows:
\begin{enumerate}
\item{Trigger Identification (TI):A trigger is correctly identified if the predicted trigger span matches with a golden label. }
\item{Trigger Classification(TC): A trigger is correctly classified if it is correctly identified and assigned to the right type.}
\item{Argument Identification (AI): An argument is correctly identified if its event type is correctly recognized and the predicted argument span matches with a golden label.}
\item{ArgumentClassification (AC): An argument is correctly classified if it is correctly identified and the predicted role matches any of the golden labels. We report Precision (P), Recall (R), and F measure (F1) for each of the four metrics.}
\end{enumerate}

Table~\ref{tab:subtasks_bioEE_models_genia} compares different performance measures of the four subtasks for different models. It could be observed that our proposed model achieves better results on some of the subtasks.
\begin{table*}[h!]
\centering
\resizebox{\textwidth}{!}
{
\begin{tabular} {|l|l|l|l|l|l|l|l|l|l|l|l|}
\textbf{Author}& Model  &Genia11 TI&Genia11 TC& Genia11 AI & Genia11 AC  & Genia11 total &Genia13 TI&Genia13 TC& Genia13 AI & Genia13 AC & Genia13 total   
\\\hline
\cite{Huang2020} &  GEANet & - & - & - & - &  60.06 & - & - & - & - & -
\\\hline
\cite{Bjorne2018} & Extended TEES & - & - & - & - &  58.10 & - & - & - & - & -
\\\hline
\cite{Zhao2021} & hypergraph & - & - & - & - &  61.10 & - & - & - & - & -
\\\hline
\cite{Li2019} & Tree-LSTM & - & - & - & - & 58.65 & - & - & - & - & -
\\\hline
\cite{FangfangSu2024} & transition-BEE & - & - & - & - & 63.24 & - & - & - & - & -
\\\hline
\cite{YingLin2020} & OneIE & - & 56.9 & - & 57.0& 56.95 & - & 57.3 & - & 51.4  & 54.35
\\\hline
\cite{ZixuanZhang2021} & AMR-IE & - & 61.5 & - & 59.8 & 60.65 & - & 63.8 & - & 58.0 & 60.9
\\\hline
\cite{JiaweiSheng2021} & CasEE & 70.0 & 67.0 &62.0 & 60.4 & 64.85 & 80.5 & 78.5 & 73.7 & 71.9 & 76.15
\\\hline
\cite{Cao2022} & OneEE &71.5& 69.5 & 65.9 & 62.5 & 67.35 & 81.9 & 80.8 & 76.8 & 72.7 & 78.05
\\\hline
ours & BioBert-GNN-BEE & 82.14 & 81.65  & 58.52  & 56.93 & 69.81 & 85.09  & 83.36  & 72.05  & 69.43 & 77.48
\\\hline
\end{tabular}
}
\caption{performance of subtasks for state-of-the-art model for biomedical event extraction for genia11 and genia13}
\label{tab:subtasks_bioEE_models_genia}
\end{table*}
Table~\ref{tab:pc_cg} compares two other datasets namely pc and cg with the baselines which shows a slight improvement.
\begin{table*}[h!]
\centering
\resizebox{\textwidth}{!}
{
\begin{tabular} 
{|l|l|l|l|l|l|l|l|l|l|l|l|}
\\\hline
\textbf{Author}& Model & cg TI & cg TC & cg AI & cg AC  & cg total  &pc11 TI & pc TC & pc AI & pc AC & pc total  
\\\hline
\cite{Miwa2015}  & EventMine  &  -  &  - & - & - & 51.33 & -  & -  & - & - & 46.97
\\\hline
\cite{Espinosa2019} & SBNN  &  -  &  - & - & - & 56.90 & -  & -  & - &  - &  -
\\\hline
\cite{Zanella2023}  & SciBERT-KG  &  82.00  &  - & 91.0 & - & - & -  & -  & -  &  -  &  -
\\\hline
\cite{LvxingZhu2020}  & hybridEE  &  -  &  - & - &  58.04 & - & -  & -  & - & - & 55.73
\\\hline
ours-RO1-withGraph & BioBert-GNN-BEE & 68.89 & 67.92  & 67.74  & 76.62  & 70.29 & 78.33  & 75.46  & 64.68  & 84.35 & 75.70
\end{tabular}
}
\caption{performance of subtasks for state-of-the-art model for biomedical event extraction for cancer genetics(cg) and path curation(pc)}
\label{tab:pc_cg}
\end{table*}
\subsection{Ablation Study}
To see the effect of using dependency parsing graph an ablation study has been carried out. The first model which is BioBertBEE does not leverage dependency parsing information while BioBertGNNBEE uses GCN to embed tokens of the sentence dependency parsing graph 
by learning a two layer GCN. 
Table~\ref{tab:ablation_bioEE_models_genia} shows the result of our ablation study which indicates that dependency parsing graph improves the performance of standard BEE.
\begin{table*}[h!]
\centering
\resizebox{\textwidth}{!}
{
\begin{tabular} {|l|l|l|l|l|l|l|l|l|l|l|l|}
\\\hline
\textbf{Author}& Model  &Genia11 TI&Genia11 TC& Genia11 AI & Genia11 AC  & Genia11 total &Genia13 TI&Genia13 TC& Genia13 AI & Genia13 AC & Genia13 total   
\\\hline
ours-RO1-noGraph & BioBert-BEE & 70.32 & 69.25  & 52.61  & 51.72 & 60.97 & 79.73  & 74.29  & 63.17  & 61.86  & 69.76
\\\hline
ours-RO1-withGraph & BioBert-GNN-BEE & 82.14 & 81.65  & 58.52  & 56.93 & 69.81 & 85.09  & 83.36  & 72.05  & 69.43 & 77.48
\\\hline
\end{tabular}
}
\caption{ablation study to see the effect of dependency parsing graph on the performance}
\label{tab:ablation_bioEE_models_genia}
\end{table*}
Similarly, Table~\ref{tab:subtasks_bioEE_models_pc_cg} shows the ablation study to see the effect of dependency parsing graph for pc and cg.
\begin{table*}[h!]
\centering
\resizebox{\textwidth}{!}
{
\begin{tabular} {|l|l|l|l|l|l|l|l|l|l|l|l|}
\\\hline
\textbf{Author}& Model & cg TI & cg TC & cg AI & cg AC  & cg total  &pc11 TI & pc TC & pc AI & pc AC & pc total  
\\\hline
ours-RO1-noGraph & BioBert-BEE & 62.83 & 61.39 & 61.28 & 59.41 & 61.22 & 70.68  & 63.91  & 68.38 & 73.19& 69.04
\\\hline
ours-RO1-withGraph & BioBert-GNN-BEE & 68.89 & 67.92  & 67.74  & 76.62  & 70.29 & 78.33  & 75.46  & 64.68  & 84.35 & 75.70
\end{tabular}
}
\caption{ablation study for biomedical event extraction for cancer genetics(cg) and path curation(pc)}
\label{tab:subtasks_bioEE_models_pc_cg}
\end{table*}
\section{Conclusion}
A new architecture is proposed based on full dependency parsing graph information. The sentence is embedded using bioBert as well as GCN. The bioBert representation only provides a contextual representation but lacks the rich structure of dependency parsing graph. To address this issue, GCN is designed on the downstream of BioBert Model to represent each token of the sentence. The head and tail networks represent the nodes of the event structure and concatenation of them provides a representation for the tasks of argument classification. These two networks also reduce model inefficiencies of GNN in a later stage of learning. Finally, An ablation study is done for all datasets to show the effect of syntactic information which may indicate that dependency parsing information provides more expressive features for BEE. 
\bibliographystyle{splncs04}
\bibliography{paper}
\end{document}